\documentclass[dvipsnames,format=sigconf,authorversion=true]{acmart}
\AtBeginDocument{%
  }

\usepackage{xcolor}
\usepackage{adjustbox}
\usepackage{colortbl}
\usepackage{tabularx}
\usepackage{booktabs} 
\usepackage{graphicx}
\usepackage[linesnumbered, vlined, ruled]{algorithm2e}
\SetArgSty{textup}
\usepackage{enumerate}
\usepackage{enumitem}
\usepackage{prettyref}
\usepackage{color, colortbl}
\usepackage{braket}
\definecolor{MyHiLiRow}{gray}{0.9}
\usepackage{tcolorbox}
\usepackage{caption}
\usepackage{seqsplit}
\usepackage{bm}
\usepackage{nameref}
\usepackage{multirow}
\usepackage[frozencache,cachedir=\detokenize{_minted}]{minted}
\usepackage{subcaption}

\usepackage{hyperref}
\usepackage{xr-hyper}
\usepackage{cleveref}

\def\vec#1{\mathbf{#1}}
\def\set#1{\mathcal{#1}}

\newcommand{\pref}{\prettyref}

\newrefformat{fig}{Figure~\ref{#1}}
\newrefformat{supfig}{Figure~\ref{#1}}
\newrefformat{tab}{Table~\ref{#1}}
\newrefformat{suptab}{Figure~\ref{#1}}
\newrefformat{sec}{Section~\ref{#1}}
\newrefformat{subsec}{Section~\ref{#1}}
\newrefformat{app}{Appendix~\ref{#1}}
\newrefformat{alg}{Algorithm~\ref{#1}}
\newrefformat{property}{Property~\ref{#1}}
\newrefformat{theorem}{Theorem~\ref{#1}}
\newrefformat{corollary}{Corollary~\ref{#1}}
\newrefformat{proposition}{Proposition~\ref{#1}}
\newrefformat{def}{Definition~\ref{#1}}
\newrefformat{eq}{equation~(\ref{#1})}

\definecolor{refpt}{RGB}{44,160,44}   
\definecolor{nearpt}{RGB}{255,127,14} 
\definecolor{solpt}{RGB}{31,119,180} 

\setcopyright{acmlicensed}

\copyrightyear{2026}
\acmYear{2026}
\setcopyright{cc}
\setcctype{by}
\acmConference[GECCO '26]{Genetic and Evolutionary Computation Conference}{July 13--17, 2026}{San Jose, Costa Rica}
\acmBooktitle{Genetic and Evolutionary Computation Conference (GECCO '26), July 13--17, 2026, San Jose, Costa Rica}
\acmDOI{10.1145/3795095.3805083}
\acmISBN{979-8-4007-2487-9/2026/07}
\begin{document}

\title[Benchmarking Stopping Criteria for Evolutionary Multi-objective Optimization]{Benchmarking Stopping Criteria for\\Evolutionary Multi-objective Optimization}


\author{Kenji Kitamura}
\affiliation{%
  \institution{Yokohama National University}
  \city{Yokohama, Kanagawa}
  \country{Japan}}
\email{kitamura-kenji-hg@ynu.jp}

\author{Ryoji Tanabe}
\affiliation{%
  \institution{Yokohama National University}
  \city{Yokohama, Kanagawa}
  \country{Japan}}
\email{tanabe-ryoji-sn@ynu.ac.jp}








\begin{abstract}


Stopping criteria automatically determine when to stop an evolutionary algorithm, so as not to waste function evaluations on a stagnant population.
Although stopping criteria play an important role in real-world applications, they have attracted little attention in the evolutionary multi-objective optimization (EMO) community.
In fact, new stopping criteria for EMO have been rarely developed in recent years.
One reason for the stagnation in developing stopping criteria for EMO is a lack of effective benchmarking methodologies.
To address this issue, this paper proposes (i) a performance measure of stopping criteria for EMO and (ii) a file-based benchmarking approach.
This paper also proposes (iii) a data representation method that effectively stores population states in text files.
(i) The proposed measure represents the performance of stopping criteria as a single scalar value, making comparison easy.
(ii) The proposed file-based approach not only simplifies the benchmarking process but also facilitates reproducibility.
(iii) The proposed data representation method addresses the issue of file size in (ii).
We demonstrate the effectiveness of our three contributions (i)--(iii) by benchmarking five representative stopping criteria for EMO.





\end{abstract}




\maketitle

\section{Introduction}
\label{sec:introduction}

\noindent \textit{General context.}
Evolutionary multi-objective optimization (EMO)~\cite{Deb01} is an effective approach for finding a non-dominated solution set that approximates the Pareto front (PF) in the objective space.
Any evolutionary algorithm has at least one termination condition; the same is true for EMO algorithms, such as NSGA-II~\cite{DebAPM02}, SMS-EMOA~\cite{beume2007sms}, and MOEA/D~\cite{ZhangL07}.

In the ideal case, an EMO algorithm should stop the search when the population perfectly approximates the PF in the objective space.
In other words, an EMO algorithm should stop the search when it finds an optimal $\mu$-distribution~\cite{AugerBBZ12} for a given quality indicator.
Unfortunately, it is impossible to determine whether a given solution set belongs to this case in real-world black-box optimization.






The most general stopping criterion is based on the maximum number of function evaluations ($\texttt{FE}^{\mathrm{max}}$) or the maximum number of iterations $t^{\mathrm{max}}$, where the search is stopped when the number of function evaluations exceeds $\texttt{FE}^{\mathrm{max}}$ or the number of iterations exceeds $t^{\mathrm{max}}$.
The drawback of this budget-based stopping criterion is the difficulty in selecting suitable $\texttt{FE}^{\mathrm{max}}$ and $t^{\mathrm{max}}$ before the search.
When $\texttt{FE}^{\mathrm{max}}$ or $t^{\mathrm{max}}$ is too small, the search is stopped early.
In this case, an EMO algorithm is likely to fail to find a good solution set.
In contrast, let us consider the case when $\texttt{FE}^{\mathrm{max}}$ or $t^{\mathrm{max}}$ is too large.
Although the EMO algorithm is likely to find a good solution set, it requires high computational cost.



Some previous studies proposed stopping criteria for EMO that automatically determine whether to stop the search.
Representative stopping criteria for EMO include   MGBM~\cite{MartiGBM07,MartiGBM16} and online convergence detection (OCD)~\cite{WagnerTN09}.
%
Throughout this paper, for simplicity, we refer to these non-budget-based stopping criteria as stopping criteria.





%




In the traditional benchmarking methodology, the performance of stopping criteria is evaluated by incorporating them into some EMO algorithms.
For example, Doush et al.~\cite{DoushEHB23} evaluated the performance of four stopping criteria by incorporating them into six EMO algorithms, including NSGA-II~\cite{DebAPM02}. 
After the EMO search is stopped, the performance of stopping criteria is evaluated from the following two aspects:
\begin{enumerate}
\renewcommand{\labelenumi}{(\arabic{enumi})}
    \item at least one quality indicator value of the final population and
    \item  the budget used in the search.
\end{enumerate}
Representative quality indicators include hypervolume (HV)~\cite{ZitzlerT98} and inverted generational distance (IGD)~\cite{CoelloS04}.
The budget is measured in terms of either the number of function evaluations (\texttt{FE}) or the number of iterations $t$.
A baseline is an EMO algorithm with the budget-based criterion that stops at $\texttt{FE}^{\mathrm{max}}$ or $t^{\mathrm{max}}$.
In the traditional benchmarking methodology, the performance of stopping criteria is evaluated in terms of (1) how much worse the quality of the final population is compared to the baseline, and (2) how much the computational budget is saved.
The results in most previous studies (e.g., ~\cite{MahbubWC15,DoushEHB23,MartiGBM07,WagnerTN09}) showed that stopping criteria can save (2) the budget while maintaining (1) the quality of the final population.



\vspace{0.5em}
\noindent \textit{Motivation.}
Although stopping criteria play a crucial role, they have not been well studied in the EMO community.
In fact, while new EMO algorithms have been frequently proposed, new stopping criteria for EMO have been rarely proposed.
Stopping criteria have also rarely been incorporated into EMO algorithms.
In fact, except for a few cases~\cite{HadkaR13}, most previous studies used the aforementioned budget-based stopping criteria.
This suggests that the performance of existing stopping criteria is insufficient for practical use.



A lack of effective benchmarking methodologies is one of the reasons for the stagnation in the development of stopping criteria for EMO.
Below, we point out two issues with the above-described traditional benchmarking methodology: 

\vspace{0.2em}
\noindent \textbf{Issue 1.}
The two aspects (1) and (2) are clearly in a trade-off relationship.
Thus, it is difficult to ``dominate" another stopping criterion in terms of both (1) and (2).
In other words, it is difficult to determine the ranking of multiple stopping criteria by taking into account (1) and (2) individually. 
This issue is discussed in \pref{subsec:issue} in detail.

\begin{figure}[t]
\centering
\includegraphics[width=0.45\textwidth]{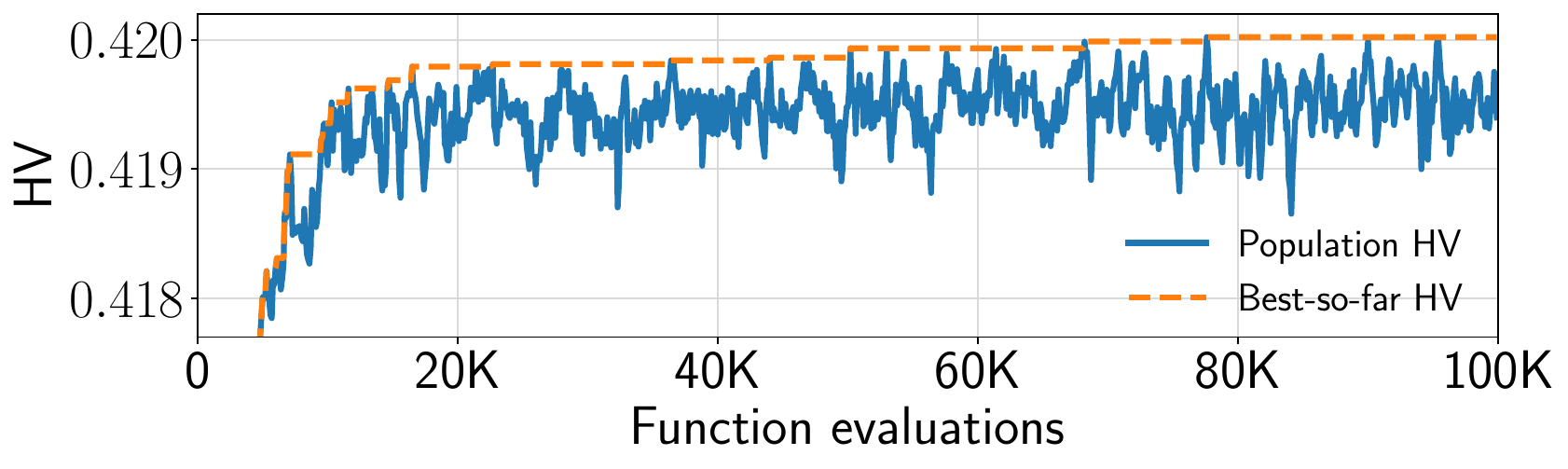}
\caption{HV values in a single run of NSGA-II.}
\label{fig:hv_transition}
\end{figure}

In addition, the HV value of a bounded set does not monotonically increase as the search progresses~\cite{Lopez-IbanezKL11}.
For example, \pref{fig:hv_transition} shows the HV values in a single run of NSGA-II on the bi-objective DTLZ2 problem. 
As shown in \pref{fig:hv_transition}, the HV value of the population does not monotonically increase.
Someone may obtain a counterintuitive observation that stopping the search early yields a better solution set than continuing the search until $\texttt{FE}^{\mathrm{max}}$.


%


\noindent \textbf{Issue 2.}
Another issue with the traditional benchmarking methodology is the difficulty in reproducing experimental results.
The traditional benchmarking methodology requires  reimplementation of the stopping criteria as well as the EMO algorithms used in their benchmarking study.
Since the behavior of EMO algorithms depends on their implementation details~\cite{Gong0PI024a,Brockhoff15}, this poses a significant obstacle to reproduction.
In addition, each time a new stopping criterion is benchmarked, the EMO algorithms must be rerun.
Such rerunning of the EMO algorithms is often difficult due to differences in computing environments, including software licenses.






\vspace{0.2em}
\noindent \textit{Contributions.}
%
%
This paper has the following three contributions (i)--(iii) that facilitate the standardization of benchmarking methodologies for stopping criteria for EMO.


\noindent \textbf{(i)} To address the first issue, this paper proposes a performance measure of stopping criteria for EMO.
Unlike the traditional approach, the proposed measure represents the two performance aspects, (1) and (2), as a single scalar.
Thus, the proposed performance measure enables a more intuitive and interpretable comparison of stopping criteria.
This is the first study to propose a performance measure for stopping criteria in the evolutionary computation community.

\noindent \textbf{(ii)} To address the second issue, this paper proposes a file-based benchmarking approach that stores population states of EMO algorithms in text files and reloads them in the benchmarking phase.
Since the proposed file-based approach does not require rerunning EMO algorithms, it simplifies the benchmarking process. 
In addition, the proposed approach improves the reproducibility.

\noindent \textbf{(iii)} However, the proposed file-based benchmarking approach raises another issue; storing the population state requires a large amount of storage.
To address this issue, this paper proposes a data representation method that significantly reduces the amount of memory required to store the population state in text files.




\vspace{0.2em}
\noindent \textit{Outline.}
\pref{sec:preliminary} introduces some preliminaries.
\pref{sec:proposed_method} presents the three contributions (i)--(iii).
\pref{sec:setup} describes our experimental setup.
\pref{sec:results} shows the results of the analysis. 
\pref{sec:conclusion} concludes this paper.


\vspace{0.2em}
\noindent \textit{Code and data availability.}
The code used in this work is available at \url{https://github.com/muzure/emo_stopping_criteria}.
Experimental data are also available at \url{https://zenodo.org/records/19435223}.

%



\section{Preliminaries}
\label{sec:preliminary}

\subsection{Multi-objective optimization}
\label{subsec:mo}

Multi-objective optimization aims to find a solution $\mathbf{x} \in \mathbb{X}$ that minimizes an objective function vector $\mathbf{f}: \mathbb{X} \rightarrow \mathbb{R}^m$, where $\mathbb{X} \subseteq \mathbb{R}^n$ is the feasible solution space, and $\mathbb{R}^m$ is the objective space.
Thus, $n$ is the dimension of the solution space, and $m$ is the dimension of the objective space.

A solution $\mathbf{x}_1$ is said to dominate $\mathbf{x}_2$ if $f_i (\mathbf{x}_1) \leq f_i (\mathbf{x}_2)$ for all $i \in \{1, \ldots, m\}$ and $f_i (\mathbf{x}_1) < f_i (\mathbf{x}_2)$ for at least one index $i$.
We denote $\mathbf{x}_1 \prec \mathbf{x}_2$ when $\mathbf{x}_1$ dominates $\mathbf{x}_2$.
In addition, $\mathbf{x}_1$ is said to weakly dominate $\mathbf{x}_2$ if $f_i (\mathbf{x}_1) \leq f_i (\mathbf{x}_2)$ for all $i \in \{1, \ldots, m\}$.
A solution $\mathbf{x}^\ast$ is a Pareto optimal solution if $\mathbf{x}^\ast$ is not dominated by any solution in $\mathbb{X}$.
The set of all Pareto optimal solutions in $\mathbb{X}$ is called the Pareto optimal solution set $\mathcal{X}^{*} = \{\mathbf{x}^* \in  \mathbb{X} \,|\, \nexists \mathbf{x} \in  \mathbb{X} \: \text{s.t.} \: \mathbf{x} \prec \mathbf{x}^* \}$.
The image of the Pareto optimal solution set in $\mathbb{R}^m$ is also called the PF $\mathbf{f}(\mathcal{X}^{*})$. 
The ideal and nadir points consist of the minimum and maximum values of the PF for $m$ objective functions, respectively.

\begin{algorithm}[t]
\small
\SetSideCommentRight
$t \leftarrow 1$, $b^{\mathrm{stop}} \leftarrow \text{False}$\;
The initialization of the population $\set{P}^{(t)} = \{\vec{x}^{(t)}_{1}, ..., \vec{x}^{(t)}_{\mu}\}$\;
\While{$b^{\mathrm{stop}} = \text{False}$, or $t < t^{\mathrm{max}}$}{
  $\set{P}'^{(t)} \leftarrow$ Mating selection ($\set{P}^{(t)}$)\;
  $\set{Q}^{(t)} \leftarrow$ Variation ($\set{P}'^{(t)}$)\;  
  $\set{P}^{(t+1)} \leftarrow $ Environmental selection$(\set{P}^{(t)} \cup \set{Q}^{(t)})$\;
  $b^{\mathrm{stop}} \leftarrow $ Stopping criterion$(\set{P}^{(t+1)}, \set{P}^{(t)}, \dots)$\;
  $t \leftarrow t + 1$\;
}
\caption{An EMO algorithm with a stopping criterion}
\label{alg:simple_emo}
\end{algorithm}

\subsection{Stopping criteria for EMO}
\label{subsec:emo_stop}

\pref{alg:simple_emo} shows a general EMO algorithm with a stopping criterion.
At the beginning of the search, the iteration counter $t$ and $b^{\mathrm{stop}}$ are initialized to 1 and false, respectively (line 1).
Here, $b^{\mathrm{stop}}$ holds a Boolean value.
If $b^{\mathrm{stop}}$ is true, the search is stopped.
Otherwise, the search continues.
After the initialization of the population $\set{P}^{(t)}$ (line 2), the following steps are repeated until either $b^{\mathrm{stop}} = $ true or $t$ reaches the maximum number of iterations $t^{\mathrm{max}}$ (lines 3--8).

At the beginning of each iteration $t$, a set of parents $\set{P}'^{(t)}$ is selected from $\set{P}^{(t)}$ by mating selection (line 4).
Then, the offspring population $\set{Q}^{(t)}$ is generated by applying genetic operators to $\set{P}'^{(t)}$ (line 5).
Individuals in the population for next iteration $\set{P}^{(t+1)}$ are selected from the union of $\set{P}^{(t)}$ and $\set{Q}^{(t)}$ by environmental selection (line 6).
At the end of each iteration $t$, the procedure of a stopping criterion is performed based on a sequence of populations $\set{P}^{(t+1)}, \set{P}^{(t)}, \dots$ (line 7).

Below, we briefly describe four representative stopping criteria that are incorporated into \pref{alg:simple_emo}.

\subsubsection{Online Convergence Detection (OCD)}
OCD~\cite{WagnerTN09} detects the convergence of EMO algorithms based on statistical results of $l$ quality indicator values.
In ~\cite{WagnerTN09}, the following three quality indicators were used: HV, the additive epsilon indicator~\cite{ZitzlerTLFF03}, and R2~\cite{HansenJ98}.
OCD stops the search when at least one of the following two conditions is satisfied during the most recent $T$ iterations:
\begin{itemize}[leftmargin=*]
  \setlength{\parskip}{0cm} 
  \setlength{\itemsep}{0cm} 
  \item The variance of $l$ quality indicator values falls below a predefined threshold $\texttt{VarLimit}$.
    \item No statistical significance is observed in $l$ quality indicator values.
\end{itemize}


At each iteration $t > T$, OCD calculates $l$ quality indicator values of $\vec{f}(\set{P}^{(t-T)}), \dots, \vec{f}(\set{P}^{(t-1)})$, where $\vec{f}(\set{P}^{(t)})$ is used as the reference point set.
Then, the $\chi^2$-variance test is applied to $\vec{I}_{t-T}, \dots, \vec{I}_{t-1}$, where $\vec{I}_{t}$ is a vector of $l$ quality indicator values of $\vec{f}(\set{P}^{(t)})$.
If the variance is smaller than $\texttt{VarLimit}$, OCD stops the search.
In addition, a regression analysis is applied to $l$ quality indicator values.
If the t-test indicates no statistically significant difference in the regression coefficients, the search is stopped.



\subsubsection{MGBM}

MGBM~\cite{MartiGBM07,MartiGBM16} was named after the initials of the authors.
For each iteration, MGBM calculates the following mutual domination rate (MDR):
\begin{align*}
    \text{MDR}\bigl(\set{P}^{(t)}, \set{P}^{(t-1)}\bigr) = \frac{|\Delta \bigl(\set{P}^{(t-1)}, \set{P}^{(t)} \bigr)|}{| \set{P}^{(t-1)} |} - \frac{|\Delta \bigl(\set{P}^{(t)}, \set{P}^{(t-1)} \bigr)|}{| \set{P}^{(t)} |},
\end{align*}
where the function $\Delta(\set{A}, \set{B})$ returns a subset of $\set{A}$ that are dominated by at least one solution in $\set{B}$.
MDR represents the improvement between two consecutive iterations $t$ and $t-1$.
It is assumed that the observed MDR values are corrupted by normally distributed noise.
To address this, based on the MDR value at each iteration, the posterior estimate is sequentially updated using a Kalman filter.
If the posterior estimate falls below a predefined threshold $\hat{I}_{\text{min}}$, MGBM stops the search.


%

\subsubsection{Entropy-based stopping criterion (ESC)}
\label{subsec:ss_entropy}

ESC~\cite{saxena2015entropy} measures a dissimilarity between two  consecutive populations $\set{P}^{(t-1)}$ and $\set{P}^{(t)}$.
First, ESC divides the objective space into a grid of $n_{\mathrm{b}}^m$ cells, where $n_{\mathrm{b}}$ is a parameter.
Then, ESC calculates the probability distribution of objective vectors in each cell.
The probability distribution for the $i$-th cell $y_i$ is given by $p(y_i) = \frac{k(y_i)}{\mu}$, where $k(y_i)$ is the number of objective vectors in $y_i$, and $\mu$ is the population size.
For each iteration $t$, ESC calculates the dissimilarity based on the probability distributions for $\set{P}^{(t-1)}$ and $\set{P}^{(t)}$.
If the dissimilarity remains unchanged for a predefined number of iterations $n_{\mathrm{s}}$, ESC stops the search. 


\subsubsection{$\epsilon$-Progress-based stopping criterion ($\epsilon$SC)}
\label{subsec:epsilon}

Borg~\cite{HadkaR13} is an EMO algorithm for many-objective optimization.
Borg uses an $\epsilon$-progress-based stopping criterion ($\epsilon$SC) to determine whether to stop the search.
Borg uses an $\epsilon$-box dominance archive that is maintained based on an $\epsilon$-box dominance relation. 
In the $\epsilon$-box dominance relation, the objective space is divided into hyper-boxes with side length $\epsilon$.
Given two objective vectors $\vec{u}=(u_1,\ldots u_m)^{\top}$ and $\vec{v}=(v_1,\ldots v_m)^{\top}$, $\vec{u}$ is said to $\epsilon$-box-dominate $\vec{v}$ if one of the following conditions is satisfied:
\begin{itemize}
  \setlength{\parskip}{0cm} 
  \setlength{\itemsep}{0cm}
    \item $\bigl\lfloor \frac{\vec{u}}{\epsilon} \bigr\rfloor \prec \bigl\lfloor \frac{\vec{v}}{\epsilon} \bigr\rfloor$
    \item $\bigl\lfloor \frac{\vec{u}}{\epsilon} \bigr\rfloor = \bigl\lfloor \frac{\vec{v}}{\epsilon} \bigr\rfloor$ and $\bigl\| \vec{u} - \bigl\lfloor \frac{\vec{u}}{\epsilon} \bigr\rfloor \bigr\| < \bigl\| \vec{v} - \bigl\lfloor \frac{\vec{v}}{\epsilon} \bigr\rfloor \bigr\|$
\end{itemize}

Let $c$ be a counter that represents the $\epsilon$-progress.
The counter $c$ is incremented if a new solution is not dominated by any solution in the $\epsilon$-box dominance archive and lies in a previously unoccupied $\epsilon$-box.
When the counter $c$ is not updated for a pre-defined number of iterations $T$, $\epsilon$SC stops the search.

\section{Proposed methods}
\label{sec:proposed_method}

This section describes the three contributions (i)--(iii) of this paper.
First, \pref{subsec:pose} proposes (i) the measure for evaluating the performance of stopping criteria for EMO.
Then, \pref{subsec:txtfile} introduces (ii) the proposed file-based benchmarking approach.
Finally, \pref{subsec:table_ascii} proposes (iii) the data representation method.

\subsection{Performance measure}
\label{subsec:pose}


This section proposes a \underline{p}erformance measure based on the \underline{o}ptimal number of function evaluations for \underline{s}topping \underline{e}volutionary algorithms (POSE).
Although any unary quality indicator can be incorporated into POSE, this section describes POSE with HV for simplicity.
Here, HV is to be maximized.



Let $\texttt{FE}^{*}$ be the optimal number of function evaluations to stop the EMO search in a single run.
$\texttt{FE}^{*}$ is a new concept introduced in this paper.
Let also $\texttt{FE}^{\mathrm{stop}}$ be the number of function evaluations when a stopping criterion actually stops the search.
The performance of the stopping criterion can be measured based on the difference between $\texttt{FE}^{\mathrm{stop}}$ and $\texttt{FE}^{*}$, which is easy to interpret.
If the difference $|\texttt{FE}^{*} -  \texttt{FE}^{\mathrm{stop}}|$ is small, it means that the stopping criterion stops the search at an appropriate point.
In contrast, if the difference $|\texttt{FE}^{*} - \texttt{FE}^{\mathrm{stop}}|$ is large, it suggests that the search is stopped too early or too late.
As discussed here, $\texttt{FE}^{*}$ provides useful information to evaluate the performance of stopping criteria.
However, it is not clear how to define $\texttt{FE}^{*}$.


Let $\texttt{bHV}^{(t)}$ be the best-so-far HV value at iteration $t$, which monotonically increases as shown in \pref{fig:hv_transition}.
Note that $\texttt{bHV}^{(t)}$ is not the HV value of the population $\set{P}^{(t)}$ at $t$. 
If $\texttt{bHV}^{(t)} - \texttt{bHV}^{(t-1)} > \delta$, we say that the best-so-far HV value is updated.
Here, $\delta \geq 0$ is a parameter that determines the update range of the HV values between two consecutive iterations.
Unless otherwise noted, $\delta = 0$.
This work defines $\texttt{FE}^{*}$ as the number of function evaluations at which the best-so-far HV value is last updated in a single run of an EMO algorithm.
Note that $\texttt{FE}^{*}$ is empirically defined a posteriori for each run, depending on the value of $\delta$.
Thus, $\texttt{FE}^{*}$ varies depending on the EMO algorithm and the problem instance.

When the search is stopped, the following cases are possible:

\begin{enumerate}[label=Case~\arabic*]
  \setlength{\parskip}{0cm}
  \setlength{\itemsep}{0cm}
\item $\texttt{FE}^{\mathrm{stop}} = \texttt{FE}^{*}$,
\item $\texttt{FE}^{\mathrm{stop}} > \texttt{FE}^{*}$, and 
\item $\texttt{FE}^{\mathrm{stop}} < \texttt{FE}^{*}$.
\end{enumerate}
Case 1 is the best case, where the search is stopped exactly at $\texttt{FE}^{*}$.
Case 2 means that the search is stopped late. 
In Case 2, $\texttt{FE}^{\mathrm{stop}} - \texttt{FE}^{*}$ function evaluations are wasted, but the quality of the solution set found by the EMO algorithm is the same as that in Case 1.
Case 3 is the opposite of Case 2.
%
In Case 3, fewer function evaluations are used than in Cases 1 and 2.
However, the quality of the solution set in Case 3 is worse than that in Cases 1 and 2.
For this reason, it is necessary to give a penalty to Case 3.

%

Based on the above discussion, this paper defines the POSE measure as follows:
\begin{align}
\label{eq:pose}
\text{POSE} =
\begin{cases}
  \frac{|\texttt{FE}^{*} - \texttt{FE}^{\mathrm{stop}}|}{\texttt{FE}^{\mathrm{max}}} & \text{if } \texttt{FE}^{\mathrm{stop}} \geq \texttt{FE}^{*} \\
  \alpha\frac{|\texttt{FE}^{*} - \texttt{FE}^{\mathrm{stop}}|}{\texttt{FE}^{\mathrm{max}}} & \text{otherwise}, 
\end{cases}
\end{align}
where the scale is normalized by the maximum number of function evaluations ($\texttt{FE}^{\mathrm{max}}$).
In \pref{eq:pose}, $\alpha \geq 1 $ is a penalty factor.
A small POSE value means that the corresponding stopping criterion stops the search at an appropriate point.
The upper part of \pref{eq:pose} represents Cases 1 and 2.
In Case 1 (i.e., $\texttt{FE}^{*} = \texttt{FE}^{\mathrm{stop}}$), the POSE value is 0, which is the minimum value.
As the stop of the search is delayed, the POSE value increases.
The lower part of \pref{eq:pose} corresponds to Case 3.
The penalty factor $\alpha$ represents the worse quality of the solution set in Case 3 than Cases 1 and 2.

In summary, POSE has two parameters: $\alpha$ and $\delta$.
We argue that the setting of $\alpha$ and $\delta$ is specified by practitioners benchmarking stopping criteria according to their preferences.
For example, if they want to impose a significant penalty on stopping criteria in Case 3, they can set $\alpha$ to a large value.
If they do not consider a small improvement of the best-so-far HV value to be meaningful, they can set $\delta$ to a non-zero value.
\pref{subsec:hyperparameter} investigates the effects of $\alpha$ and $\delta$.



\subsection{File-based benchmarking approach}
\label{subsec:txtfile}

As pointed out in \pref{sec:introduction}, the traditional benchmarking methodology requires reimplementing EMO algorithms and rerunning them.
These not only hinder reproduction but also make the benchmarking process computationally expensive.
To address this issue, we propose a file-based benchmarking approach.




As described in \pref{subsec:emo_stop}, stopping criteria determine whether to stop the search based on the state of the population $\set{P}$.
Without exception, all existing stopping criteria for EMO use the objective vector set of the population $\vec{f}(\set{P})$.
For simplicity, the population state refers to $\vec{f}(\set{P})$ throughout this paper.
Two important facts are as follows:
\begin{itemize}
  \setlength{\parskip}{0cm} 
  \setlength{\itemsep}{0cm}
    \item Stopping criteria do not influence the behavior of EMO algorithms until the search is stopped, and
    \item Stopping criteria rely solely on the population states up to that point.
\end{itemize}

Suppose that all objective vector sets of the populations $\vec{f}(\set{P}^{(1)}),$ $\dots, \vec{f}(\set{P}^{(t^{\mathrm{max}})})$ for all $t^{\mathrm{max}}$ iterations are stored in $t^{\mathrm{max}}$ text files $\texttt{fP\_1.csv}, $ $\dots, $ $\texttt{fP\_tmax.csv}$, respectively.
Here, $t^{\mathrm{max}}$ is the maximum number of iterations.
\pref{fig:example_naive} shows examples of these text files for bi-objective optimization, where the population size $\mu=4$, the offspring population size $\lambda=1$, and $t^{\mathrm{max}}=3$.
Bold font in \pref{fig:example_naive} indicates the difference between $\vec{f}(\set{P}^{(t-1)})$ and $\vec{f}(\set{P}^{(t)})$.

Let us consider traditional benchmarking of a stopping criterion (e.g., OCD) by incorporating it into an EMO algorithm (e.g., NSGA-II).
Based on a sequence of population states $\vec{f}(\set{P}^{(1)}), \dots, \vec{f}(\set{P}^{(t)})$, the stopping criterion determines whether to stop the search at iteration $t$.
We point out that this process can be perfectly simulated by sequentially reloading the above-described text files $\texttt{fp\_1.csv}, $ $\dots, $ $\texttt{fp\_t.csv}$ and giving them to the stopping criterion.
This paper refers to this approach as the file-based benchmarking approach.

Once EMO algorithms are executed and their population states are stored in files, the file-based benchmarking approach does not require rerunning the EMO algorithms.
%
%
Since the proposed approach does not require reimplementing EMO algorithms, it also improves the reproducibility~\cite{Lopez-IbanezBP21}.
If the text files are publicly available on the Internet as in this paper, researchers around the world can reproduce the experimental results of stopping criteria for EMO.




\subsection{Data representation method}
\label{subsec:table_ascii}


\begin{figure}[t]
\centering
\begin{minipage}[t]{0.11\textwidth}
\begin{minted}[frame=single,bgcolor=gray!10,linenos,xleftmargin=1em,numbersep=0.1em]{text}
1.78,2.53
3.14,2.91
0.26,4.55
2.88,0.98
\end{minted}
\subcaption{$\texttt{fP\_1.csv}$}
\label{fig:p1}
\end{minipage}
\hspace{0.04\columnwidth}
\begin{minipage}[t]{0.11\textwidth}
\begin{minted}[frame=single,bgcolor=gray!10,linenos,xleftmargin=1em,numbersep=0.1em,escapeinside=||]{text}
1.78,2.53
3.14,2.91
|\textbf{1.27,2.55}|
2.88,0.98
\end{minted}
\subcaption{$\texttt{fP\_2.csv}$}
\label{fig:p2}
\end{minipage}
\hspace{0.04\columnwidth}
\begin{minipage}[t]{0.11\textwidth}
\begin{minted}[frame=single,bgcolor=gray!10,linenos,xleftmargin=1em,numbersep=0.1em,escapeinside=||]{text}
|\textbf{1.45,2.39}|
3.14,2.91
1.27,2.55
2.88,0.98
\end{minted}
\subcaption{$\texttt{fP\_3.csv}$}
\label{fig:p3}
\end{minipage}
\caption{Examples of three text files that maintain the three objective vector sets $\vec{f}(\set{P}^{(1)}), \vec{f}(\set{P}^{(2)}),$ and $\vec{f}(\set{P}^{(3)})$, respectively.}
\label{fig:example_naive}
\end{figure}

\begin{figure}[t]
\centering

\begin{minipage}[t]{0.11\textwidth}
\begin{minted}[frame=single,bgcolor=gray!10,linenos,xleftmargin=1em,xrightmargin=0pt, framesep=1pt,numbersep=0.1em]{text}
1.78,2.53
3.14,2.91
0.26,4.55
2.88,0.98
1.27,2.55
1.45,2.39
\end{minted}
\subcaption{\texttt{fx.csv}}
\label{fig:fx}
\end{minipage}
\hspace{0.04\columnwidth}
\begin{minipage}[t]{0.1\textwidth}
\begin{minted}[frame=single,bgcolor=green!10,linenos,xleftmargin=1em,numbersep=0.1em]{text}
1,2,3,4
1,2,5,4
6,2,5,4
\end{minted}
\subcaption{\texttt{id.csv}}
\label{fig:fx}
\end{minipage}
\caption{Examples of the proposed method.}
\label{fig:example_proposed}
\end{figure}


The simple data representation method described in \pref{subsec:txtfile} stores $t^{\mathrm{max}}$ objective vector sets $\vec{f}(\set{P}^{(1)}),$ $\dots, \vec{f}(\set{P}^{(t^{\mathrm{max}})})$ in $t^{\mathrm{max}}$ text files $\texttt{fP\_1.csv}, $ $\dots, $ $\texttt{fP\_tmax.csv}$, respectively.
Since this method needs to store $m \times \mu \times t^{\mathrm{max}}$ real-valued objective vectors in text files, it requires a large amount of storage.
As demonstrated in \pref{subsec:fsize}, this is especially problematic for steady-state EMO algorithms (e.g., SMS-EMOA) and many-objective optimization.
As discussed in \pref{subsec:txtfile}, the text files are expected to be uploaded to a public repository; therefore, an efficient data representation method is required to reduce file size.


To address this issue, this section proposes a data representation method for effectively storing population states in text files.
We point out that the difference between two consecutive populations $\set{P}^{(t-1)}$ and $\set{P}^{(t)}$ in elitist EMO algorithms is generally not significant.
This is because some individuals of $\set{P}^{(t-1)}$ survive in the next population $\set{P}^{(t)}$. 
This is especially true just before the search stagnates. 
For example, the difference between $\vec{f}(\set{P}^{(1)})$ and $\vec{f}(\set{P}^{(2)})$ in \pref{fig:example_naive} is only the third objective vector, where it is the objective vector of a child at $t=2$, and $\vec{f}(\vec{x}_3)$ in $\vec{f}(\set{P}^{(1)})$ is replaced by it.
It is possible to reduce file size by not storing individuals that have already appeared in earlier populations.
%

The proposed method uses the following two files:
\begin{description}
  \setlength{\parskip}{0cm} 
  \setlength{\itemsep}{0cm} 
    \item[\texttt{fx.csv}] that maintains objective vectors of all $\texttt{FE}^{\mathrm{max}}$ individuals, and 
    \item[\texttt{id.csv}] that maintains IDs of $\mu$ individuals in the population for each iteration.
\end{description}
Here, $\texttt{FE}^{\mathrm{max}}$ is the maximum number of evaluations, where $\texttt{FE}^{\mathrm{max}}=\mu + \lambda \times (t^{\mathrm{max}}-1)$ in general EMO algorithms.
\pref{fig:example_proposed} shows examples of \texttt{fx.csv} and \texttt{id.csv}, where the setting of $m$, $\mu$, $\lambda$, and $t^{\mathrm{max}}$ is the same as in \pref{fig:example_naive}.
In this case, $\texttt{FE}^{\mathrm{max}}=6$, where $4+1 \times 2 = 6$.
As shown in \pref{fig:example_proposed}(a), \texttt{fx.csv} also stores $\texttt{FE}^{\mathrm{max}}$ objective vectors, one per line; that is, it contains $\texttt{FE}^{\mathrm{max}}$ lines.
As shown in \pref{fig:example_proposed}(b), \texttt{id.csv} stores $\mu$ IDs, one per line; that is, it contains $t^{\mathrm{max}}$ lines.
An ID in \texttt{id.csv} corresponds to the line number in \texttt{fx.csv}.






The objective vector set of the population $\vec{f}(\set{P}^{(t)})$ at iteration $t$ can be easily reproduced by the following two steps, where $\set{F}$ represents the objective vector set $\vec{f}(\set{P}^{(t)})$, and $\set{F} = \emptyset$ at first:
\begin{enumerate}[label=Step~\arabic*]
  \setlength{\parskip}{0cm}
  \setlength{\itemsep}{0cm}
\item Obtain a set of $\mu$ IDs $\set{I} = \{i_1, \dots, i_{\mu}\}$ by loading the $t$-th line in \texttt{id.csv}.
\item For each $i \in \set{I}$, select the objective vector at the $i$-th line in \texttt{fx.csv} and add it to $\set{F}$.
\end{enumerate}
In the example of \pref{fig:example_proposed}, $\vec{f}(\set{P}^{(1)})$ at $t=1$ is reproduced by selecting the four objective vectors at lines 1, 2, 3, and 4 in \texttt{fx.csv}.
Similarly, $\vec{f}(\set{P}^{(2)})$ at $t=2$ is reproduced by selecting the four objective vectors at lines 1, 2, 5, and 4 in \texttt{fx.csv}.
The same applies to $\vec{f}(\set{P}^{(3)})$ at $t=3$.
The resulting $\vec{f}(\set{P}^{(1)})$, $\vec{f}(\set{P}^{(2)})$, and $\vec{f}(\set{P}^{(3)})$ are exactly the same as those in  $\texttt{fP\_1.csv}$, $\texttt{fP\_2.csv}$, and  $\texttt{fP\_3.csv}$ in \pref{fig:example_naive}, respectively.
Unlike $\texttt{fP\_1.csv}$, $\texttt{fP\_2.csv}$, and  $\texttt{fP\_3.csv}$ in \pref{fig:example_naive}, the proposed data representation method does not store objective vectors that have already appeared in past populations.
This significantly reduces the amount of memory required to store the population state in text files.

The proposed data representation method can handle solutions in the same manner as objective vectors.
Here, only one stopping criterion~\cite{MahbubWC15} uses the population state in the solution space (i.e., $\set{P}$) in addition to the objective space.
One stopping criterion~\cite{GoelN10}) is based on an unbounded external archive that maintains all non-dominated solutions found so far by an EMO algorithm.
The proposed file-based approach can reproduce the unbounded archive $\set{A}^{(t)}$ at iteration $t$ by simply loading lines 1 through $(\mu + \lambda \times (t-1))$ from \texttt{fx.csv} and selecting non-dominated objective vectors from them.


\section{Experimental setup}
\label{sec:setup}

This section describes the experimental setup for our analysis.
We evaluated the performance of the four stopping criteria (OCD, MGBM, ESC, and $\epsilon$SC) described in \pref{subsec:emo_stop}.
In addition, as a baseline, this work considers a simple stopping criterion that stops the search when the best-so-far HV is not updated for $T$ consecutive iterations.
We refer to this method as an indicator-based stopping criterion (ISC).
All five stopping criteria were implemented in Python.

Table \ref{tab:supp-sc_pram} in the supplementary file shows the parameters of the five stopping criteria.
Except for $\hat{I}_{\min}$ in MGBM, the parameters were set to the default values.
In our preliminary experiment, we observed that MGBM with the default $\hat{I}_{\min}$ value ($0.0001$) does not stop the search in any case.
For this reason, $\hat{I}_{\min}$ was set to $0.12$ so that MGBM stops the search before reaching the maximum number of evaluations.


We incorporated the five stopping criteria into the following five representative EMO algorithms: NSGA-II~\cite{DebAPM02}, SMS-EMOA~\cite{beume2007sms}, IBEA~\cite{ZitzlerK04}, MOEA/D~\cite{ZhangL07}, and NSGA-III~\cite{DebJ14}.
We used \texttt{pymoo}~\cite{BlankD20} implementations of the five EMO algorithms. 
The population size was set to $100$.
The maximum number of function evaluations was set to $10^5$.
For each problem, 31 independent runs were performed.

We used the DTLZ1, ..., DTLZ7 problems~\cite{deb2002scalable} and the convex version of DTLZ2 (CDTLZ2) \cite{DebJ14}.
The number of objectives $m$ was set as follows: $m \in \{2, 4, 6\}$.
We used HV~\cite{ZitzlerT98} as a quality indicator.
For HV calculation, first, all objective vectors are normalized into $[0,1]^m$ by using the true ideal and nadir points of each problem.
Then, the reference point was set to $(1.1, \dots, 1.1)^{\top}$.
The \texttt{pygmo}~\cite{biscani2020parallel} code was used for HV calculation.







\definecolor{c1}{RGB}{150,150,150}
\definecolor{c2}{RGB}{220,220,220}

\section{Results}
\label{sec:results}

This section shows our analysis results.
\pref{subsec:issue} demonstrates the first issue of the traditional performance evaluation method pointed out in \pref{sec:introduction}.
Then, \pref{subsec:eff_pose} investigates the effectiveness of the proposed POSE.
\pref{subsec:hyperparameter} investigates the impact of the two parameters $\alpha$ and $\delta$ on POSE.
%
Finally, \pref{subsec:fsize} shows that the proposed data representation method can effectively reduce file size.

Note that a rigorous benchmarking of the five stopping criteria is beyond the scope of this paper.
Our preliminary results show that the performance of the five stopping criteria is highly sensitive to their parameter settings.
However, parameter tuning for stopping criteria is not straightforward; moreover, no previous studies have addressed this issue.

\subsection{Drawback of the traditional performance evaluation method}
\label{subsec:issue}

\begin{figure}[t]
\centering
\includegraphics[width=0.45\textwidth]{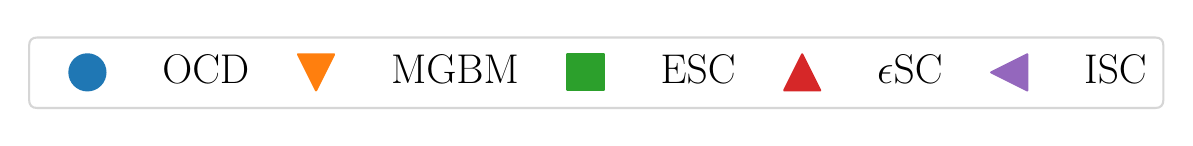}\\[-0.7em]
\subfloat[DTLZ1]{
\includegraphics[width=0.225\textwidth]{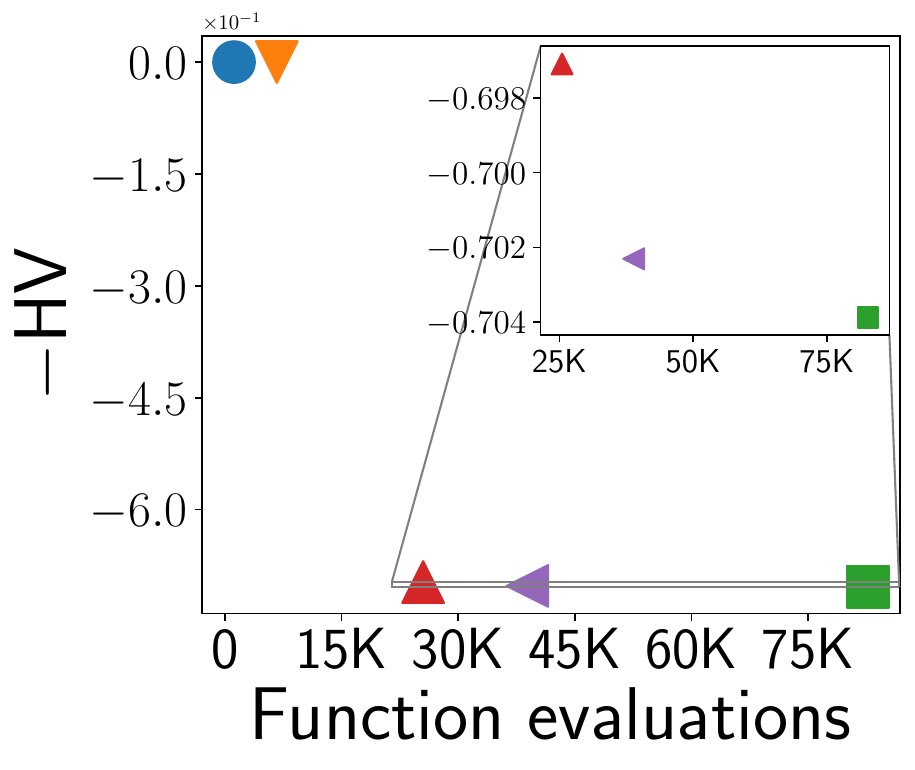}
}
\subfloat[DTLZ2]{
\includegraphics[width=0.225\textwidth]{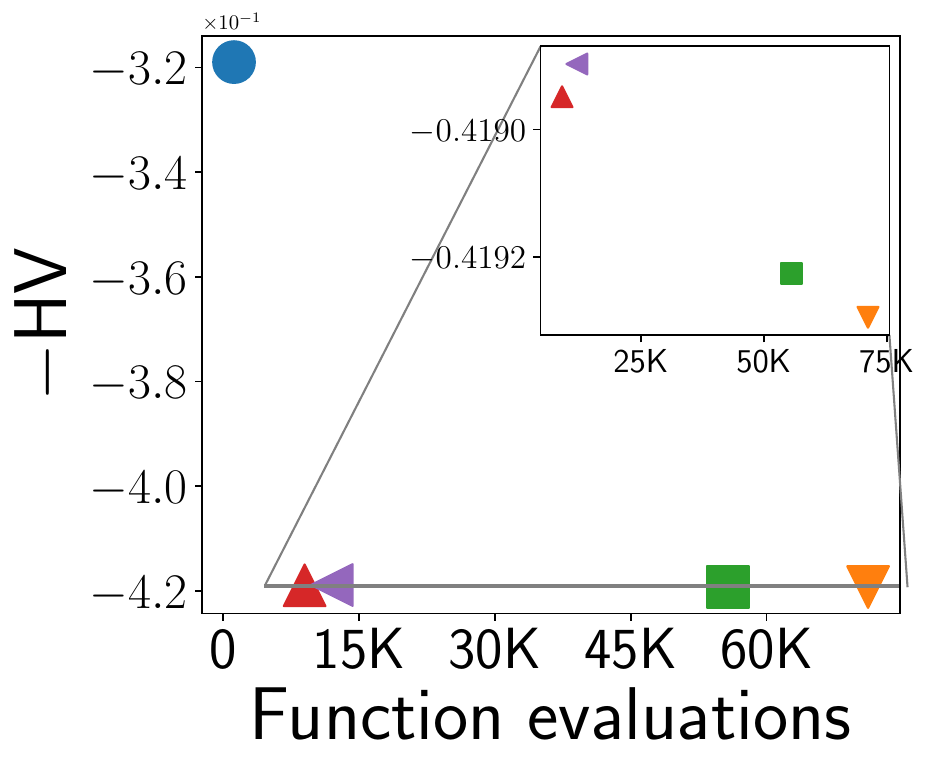}
}
\caption{Benchmarking results by the traditional method.}
  \label{fig:comparison of strategies}
\end{figure}

\pref{fig:comparison of strategies} shows the results of NSGA-II with the five stopping criteria on the bi-objective DTLZ1 and DTLZ2 problems in a single run.
In \pref{fig:comparison of strategies}, the vertical axis represents (1) the HV values of the final populations, and the horizontal axis represents (2) the number of function evaluations used in the search.
For the sake of readability, $-$HV values are shown in \pref{fig:comparison of strategies}. 
Since this experiment was conducted by the proposed file-based approach described in \pref{subsec:txtfile}, the behavior of NSGA-II is exactly the same until the end of the search.
Figure \ref{fig:supp-traditional_approach} in the supplementary file shows the results on all problems with $m \in \{2, 4, 6\}$.
Due to the space limitation, we do not describe them, but most of them are similar to \pref{fig:comparison of strategies}.

Below, we evaluate the performance of the five stopping criteria based on the traditional method described in \pref{sec:introduction}.
Thus, their performance is independently evaluated from the two aspects: (1) the quality of the final population and (2) the budget used in the search.


As shown in \pref{fig:comparison of strategies}(a), ESC performs the best in terms of (1), followed by ISC.
The worst performers are OCD and MGBM in terms of (1).
In contrast, OCD performs the best in terms of (2), followed by MGBM.
ESC achieves the worst performance in terms of (2).
In summary, OCD and ESC are \textit{non-dominated} with respect to each other under the two conflicting aspects (1) and (2).
The same relation can be observed in the other stopping criteria.
For example, in the comparison between ESC and $\epsilon$SC, ESC outperforms $\epsilon$SC in terms of (1) but is outperformed by $\epsilon$SC in terms of (2).

The same is true for the results on the DTLZ2 problem shown in \pref{fig:comparison of strategies}(b).
OCD, MGBM, ESC, and $\epsilon$SC are non-dominated with each other, and there is no clear winner among them.
The comparative result between ISC and $\epsilon$SC is counterintuitive.
Although ISC uses more function evaluations than $\epsilon$SC, its HV value is worse.
This is due to the lack of monotonicity~\cite{Lopez-IbanezKL11}   described in \pref{sec:introduction}.


\vspace{0.2em}
\noindent \textit{Summary.} Our results demonstrate the difficulty of evaluating the performance of stopping criteria independently in terms of the two aspects (1) and (2).

\subsection{Effectiveness of POSE}
\label{subsec:eff_pose}



\begin{figure}[t]
\centering
\includegraphics[width=0.45\textwidth]{figs/legend_comp_sc.pdf}\\[-0.5em]
\includegraphics[width=0.45\textwidth]{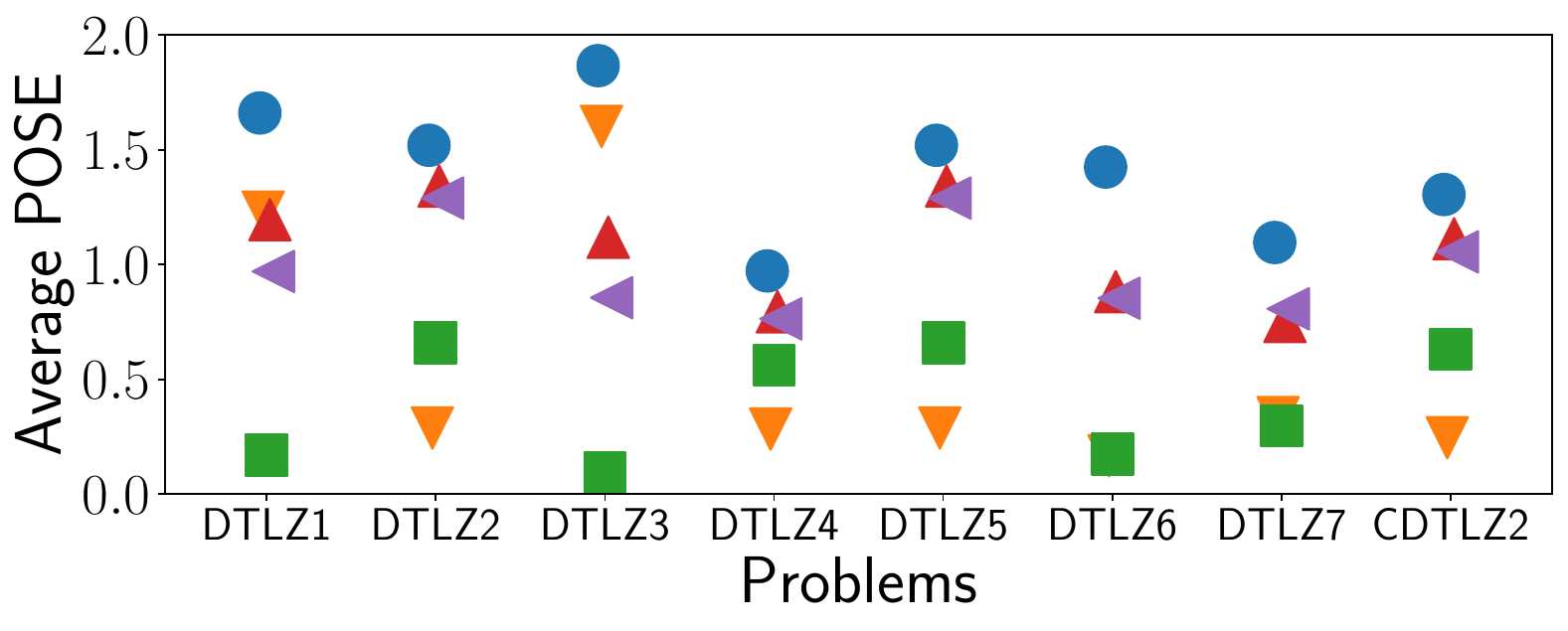}
\caption{Distributions of the average POSE values.} 
  \label{fig:pose}
%
\includegraphics[width=0.45\textwidth]{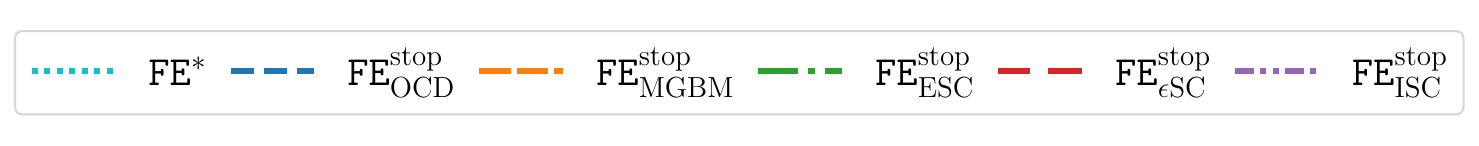}\\[-0.1em]
\includegraphics[width=0.35\textwidth]{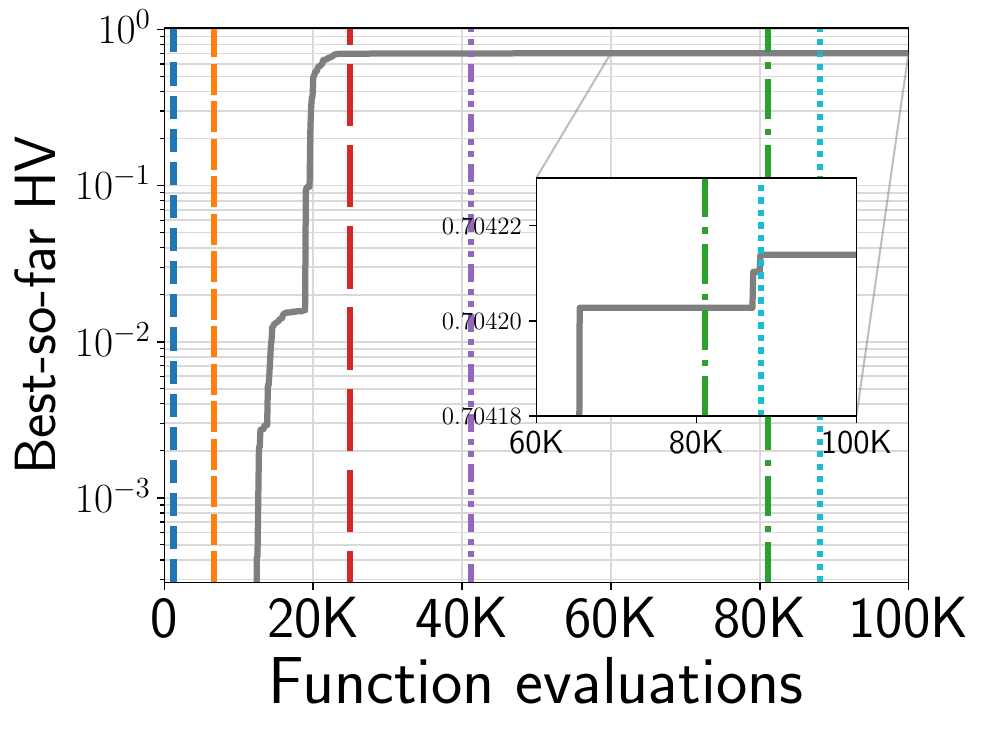}
\caption{$\texttt{FE}^{*}$ and $\texttt{FE}^{\mathrm{stop}}$ of the five stopping criteria on the DTLZ1 problem with $m=2$. The gray line represents the best-so-far HV value.}
  \label{fig:fe_star and fe_stop}
\end{figure}

\pref{fig:pose} shows the results of the five stopping criteria on the eight problems, where they were incorporated into NSGA-II.
\pref{fig:pose} shows the average POSE values over 31 runs for each stopping criterion and problem.
Here, a small POSE value indicates a better performance of stopping criteria.
In this section, $\alpha$ and $\delta$ in POSE are set to $2$ and $0$, respectively.
Table \ref{tab:supp-pose} (a) in the supplementary file is a tabular representation of \pref{fig:pose}.

As seen from \pref{fig:pose}, MGBM and ESC perform well in most cases.
MGBM shows the best performance on the DTLZ2, DTLZ4, DTLZ5, DTLZ6, and CDTLZ2 problems.
ESC also performs the best on the DTLZ1, DTLZ3, and DTLZ7 problems.
Here, no statistically significant difference is observed between the results of MGBM and ESC on the DTLZ6 problem according to the Wilcoxon rank-sum test.
In contrast, OCD achieves the worst performance on all problems.


\pref{fig:fe_star and fe_stop} shows $\texttt{FE}^*$ and $\texttt{FE}^{\mathrm{stop}}$ of the five stopping criteria on the bi-objective DTLZ1 problem in a single run.
The gray line in \pref{fig:fe_star and fe_stop} shows the best-so-far HV value at each iteration.
Figure \ref{fig:supp-fe_star and fe_stop} in the supplementary file shows the results on all problems, but it is similar to \pref{fig:fe_star and fe_stop}.


As shown in \pref{fig:fe_star and fe_stop}, MGBM and ESC stop the search once the improvement in the best HV value has sufficiently converged.
In contrast, OCD tends to stop the search in the early stage of the search, where the HV value increases rapidly.
Since the update range $\delta$ in POSE is set to $0$ in this experiment, $\texttt{FE}^*$ is likely to be the later stage of the search.
This mismatch is one of the reasons why OCD performs poorly in \pref{fig:fe_star and fe_stop}.
In fact, the results in \pref{subsec:hyperparameter} show that OCD achieves a better POSE value when using a non-zero $\delta$ value.

Note that the ranking of the five stopping criteria depends on the type of EMO algorithm. 
Figures \ref{fig:supp-pose_avg_nsga2}--\ref{fig:supp-pose_avg_nsga3} in the supplementary file show the results when using NSGA-II, IBEA, SMS-EMOA, MOEA/D, and NSGA-III for $m \in \{2,4,6\}$.
We do not describe Figures \ref{fig:supp-pose_avg_nsga2}--\ref{fig:supp-pose_avg_nsga3} in detail, but they show that the best stopping criterion varies depending on the type of EMO algorithm.

\begin{figure}[t]
\centering
\includegraphics[width=0.45\textwidth]{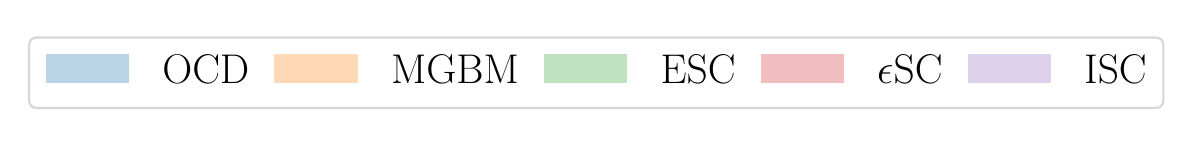}\\[-0.7em]
\centering
\subfloat[DTLZ1]{
\includegraphics[width=0.22\textwidth]{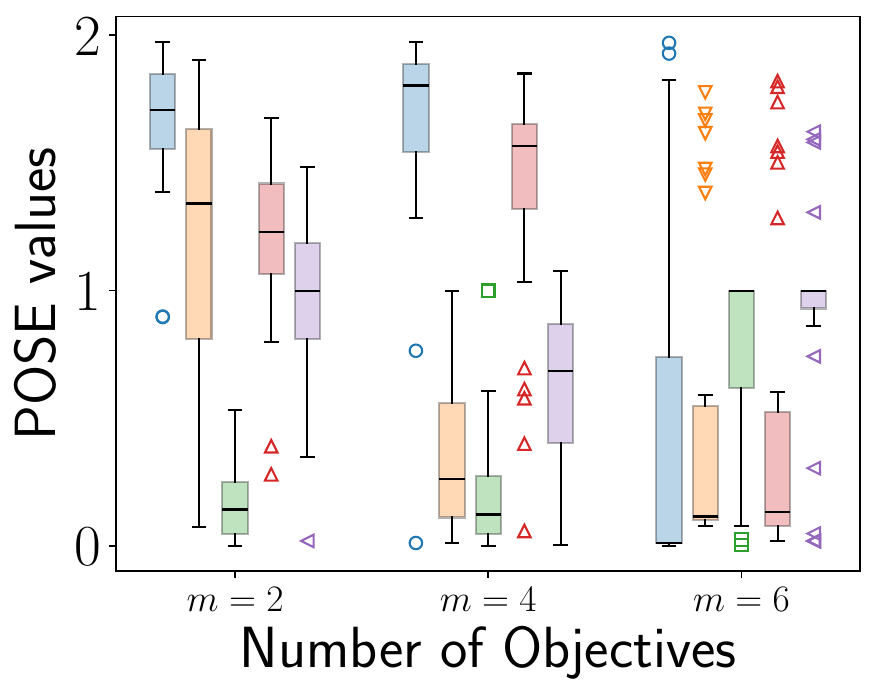}
}
\subfloat[DTLZ2]{
\includegraphics[width=0.22\textwidth]{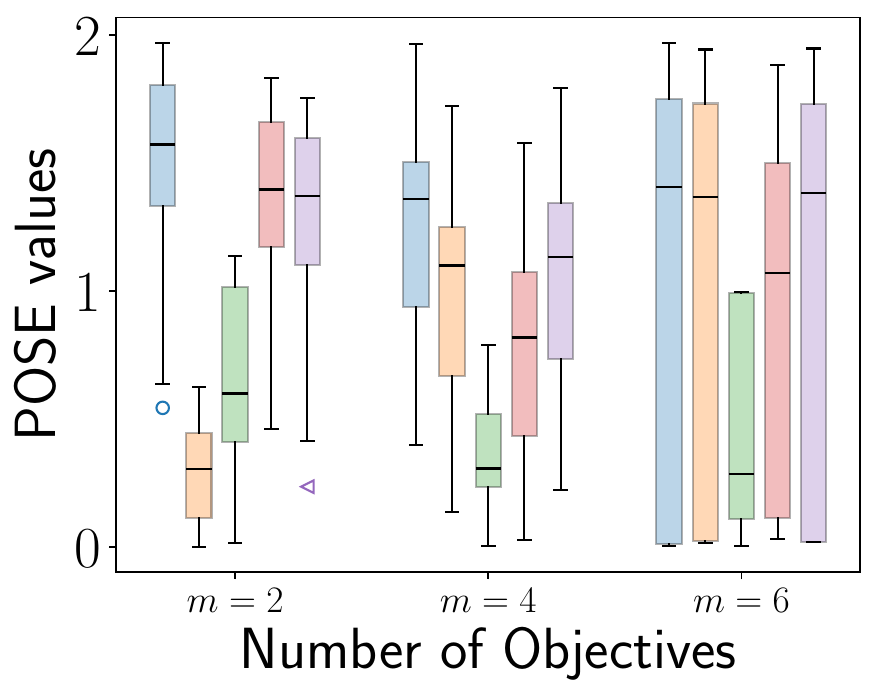}
}
\caption{Distributions of POSE values for $m \in \{2, 4, 6\}$ when using NSGA-II.}
\label{fig:pose_n_obj_comp}
\end{figure}

\vspace{0.2em}
\noindent \textit{Scalability to the number of objectives $m$.} 
\pref{fig:pose_n_obj_comp} shows the results of the five stopping criteria on the DTLZ1 and DTLZ2 problems with $m=\{2,4,6\}$ when using NSGA-II.
Each box in \pref{fig:pose_n_obj_comp} represents the distribution of 31 POSE values for each EMO algorithm.
Figures \ref{fig:supp-pose_n_obj_comp_nsga2}--\ref{fig:supp-pose_n_obj_comp_nsga3} show the results of all five EMO algorithms on all eight problems.

As shown in \pref{fig:pose_n_obj_comp}, the distributions of the POSE values depend on $m$ and the type of problem.
For example, as seen from \pref{fig:pose_n_obj_comp}(a), OCD achieves the worst POSE values on the DTLZ1 problem with $m=2, 4$.
In contrast, OCD performs the best for $m=6$.
As seen from \pref{fig:pose_n_obj_comp}(b), the width of the boxes increases on the DTLZ2 problem as $m$ increases.
Similar observations as in \pref{fig:pose_n_obj_comp}(b) can be drawn from Figures \ref{fig:supp-pose_n_obj_comp_nsga2}--\ref{fig:supp-pose_n_obj_comp_nsga3} in most cases.
This means that the stopping criteria exhibit a large variation in their performance across runs as $m$ increases.


\vspace{0.2em}
\noindent \textit{Summary.} The results demonstrate that the proposed POSE enables a straightforward comparison of stopping criteria for EMO while considering both aspects (1) and (2), even for large values of $m$.

\begin{table}[t]
\centering%
\caption{Friedman test-based average rankings of the five stopping criteria for each $\alpha$, where $\delta=0$. 
}
\label{tab:pose_param_alpha}
{\small
\subfloat[$m=2$]{
\begin{tabular}{lccccc}
\toprule
& $\alpha=1$ &$\alpha=2$ & $\alpha=3$ & $\alpha=4$ & $\alpha=5$\\
\midrule
OCD & 5.00 & 5.00 & 5.00 & 5.00 & 5.00\\
MGBM & \cellcolor{c1}1.75 & \cellcolor{c2}1.88 & \cellcolor{c2}2.00 & \cellcolor{c2}2.00 & \cellcolor{c2}2.00\\
ESC & \cellcolor{c2}1.75 & \cellcolor{c1}1.62 & \cellcolor{c1}1.50 & \cellcolor{c1}1.50 & \cellcolor{c1}1.50\\
$\epsilon$SC & 3.62 & 3.62 & 3.62 & 3.62 & 3.62\\
ISC & 2.88 & 2.88 & 2.88 & 2.88 & 2.88\\
\bottomrule
\end{tabular}
}
\\
\subfloat[$m=4$]{
\begin{tabular}{lccccc}
\toprule
& $\alpha=1$ &$\alpha=2$ & $\alpha=3$ & $\alpha=4$ & $\alpha=5$\\
\midrule
OCD & 3.88 & 4.00 & 4.00 & 4.00 & 4.00\\
MGBM & \cellcolor{c2}2.50 & 2.75 & \cellcolor{c2}2.62 & \cellcolor{c2}2.62 & \cellcolor{c2}2.62\\
ESC & 2.56 & \cellcolor{c1}1.81 & \cellcolor{c1}1.94 & \cellcolor{c1}1.94 & \cellcolor{c1}1.94\\
$\epsilon$SC & \cellcolor{c1}2.38 & \cellcolor{c2}2.62 & 2.62 & 2.62 & 2.62\\
ISC & 3.69 & 3.81 & 3.81 & 3.81 & 3.81\\
\bottomrule
\end{tabular}
}
\\
\subfloat[$m=6$]{
\begin{tabular}{lccccc}
\toprule
& $\alpha=1$ &$\alpha=2$ & $\alpha=3$ & $\alpha=4$ & $\alpha=5$\\
\midrule
OCD & \cellcolor{c1}2.50 & \cellcolor{c1}2.50 & \cellcolor{c2}2.75 & \cellcolor{c2}2.88 & \cellcolor{c2}2.88\\
MGBM & 2.88 & 2.88 & 2.75 & 2.88 & 2.88\\
ESC & 3.56 & 3.31 & 3.31 & 2.94 & 2.94\\
$\epsilon$SC & \cellcolor{c2}2.50 & \cellcolor{c2}2.62 & \cellcolor{c1}2.50 & \cellcolor{c1}2.62 & \cellcolor{c1}2.62\\
ISC & 3.56 & 3.69 & 3.69 & 3.69 & 3.69\\
\bottomrule
\end{tabular}
}
}
\end{table}

\begin{table}[t]
\centering%
\caption{Friedman test-based average rankings of the five stopping criteria for each $\delta$, where $\alpha=2$. 
}
\label{tab:pose_param_delta}
{\small
\subfloat[$m=2$]{
\begin{tabular}{lccccc}
\toprule
& $\delta=0$ &$\delta=10^{-4}$ & $\delta=10^{-3}$ & $\delta=10^{-2}$ & $\delta=10^{-1}$\\
\midrule
OCD & 5.00 & 4.62 & \cellcolor{c2}2.12 & \cellcolor{c2}2.00 & \cellcolor{c1}1.38\\
MGBM & \cellcolor{c2}1.88 & 3.25 & 4.12 & 4.25 & 4.12\\
ESC & \cellcolor{c1}1.62 & \cellcolor{c2}2.38 & 4.50 & 4.50 & 4.50\\
$\epsilon$SC & 3.62 & 2.88 & \cellcolor{c1}1.88 & \cellcolor{c1}1.88 & \cellcolor{c2}2.25\\
ISC & 2.88 & \cellcolor{c1}1.88 & 2.38 & 2.38 & 2.75\\
\bottomrule
\end{tabular}
}
\\
\subfloat[$m=4$]{
\begin{tabular}{lccccc}
\toprule
& $\delta=0$ &$\delta=10^{-4}$ & $\delta=10^{-3}$ & $\delta=10^{-2}$ & $\delta=10^{-1}$\\
\midrule
OCD & 4.00 & 4.00 & 3.88 & \cellcolor{c1}2.50 & \cellcolor{c1}1.50\\
MGBM & 2.75 & 2.75 & \cellcolor{c2}2.50 & 2.62 & 3.12\\
ESC & \cellcolor{c1}1.81 & \cellcolor{c1}1.81 & 2.69 & 4.31 & 4.44\\
$\epsilon$SC & \cellcolor{c2}2.62 & \cellcolor{c2}2.62 & \cellcolor{c1}2.38 & \cellcolor{c2}2.50 & 3.38\\
ISC & 3.81 & 3.81 & 3.56 & 3.06 & \cellcolor{c2}2.56\\
\bottomrule
\end{tabular}
}
\\
\subfloat[$m=6$]{
\begin{tabular}{lccccc}
\toprule
& $\delta=0$ &$\delta=10^{-4}$ & $\delta=10^{-3}$ & $\delta=10^{-2}$ & $\delta=10^{-1}$\\
\midrule
OCD & \cellcolor{c1}2.50 & \cellcolor{c1}2.50 & \cellcolor{c1}2.12 & \cellcolor{c1}1.62 & \cellcolor{c1}1.00\\
MGBM & 2.88 & 2.88 & \cellcolor{c2}2.50 & \cellcolor{c2}2.25 & \cellcolor{c2}2.38\\
ESC & 3.31 & 3.31 & 4.19 & 4.19 & 4.69\\
$\epsilon$SC & \cellcolor{c2}2.62 & \cellcolor{c2}2.62 & 2.88 & 3.25 & 3.50\\
ISC & 3.69 & 3.69 & 3.31 & 3.69 & 3.44\\
\bottomrule
\end{tabular}
}
}
\end{table}

\subsection{Effects of $\alpha$ and $\delta$ in POSE}
\label{subsec:hyperparameter}




This section analyzes the effects of $\alpha$ and $\delta$ in POSE in order to provide a rule of thumb for setting them.

\subsubsection{Effects of $\alpha$}

\pref{tab:pose_param_alpha} shows the results of the five stopping criteria on all eight DTLZ problems with $m \in \{2, 4, 6\}$ when using NSGA-II.
\pref{tab:pose_param_alpha} shows the Friedman test-based average rankings of the five stopping criteria when setting $\alpha$ to $1, 2, 3, 4,$ and $5$, where $\delta=0$ in this analysis.
The \texttt{CONTROLTEST} software~\cite{GarciaFLH10} (\url{https://sci2s.ugr.es/sicidm}) was used to calculate the rankings.
In \pref{tab:pose_param_alpha}, the best and second-best data are highlighted in {\adjustbox{margin=0.1em, bgcolor=c1}{dark gray} and {\adjustbox{margin=0.1em, bgcolor=c2}{gray}, respectively.
Tables \ref{tab:pose_param_alpha_IBEA}--\ref{tab:pose_param_alpha_NSGA3} in the supplementary file show the results of the other EMO algorithms.
Here, the same conclusion as in \pref{tab:pose_param_alpha} can be drawn from Tables \ref{tab:pose_param_alpha_IBEA}--\ref{tab:pose_param_alpha_NSGA3}.

As shown in \pref{tab:pose_param_alpha}, for $\alpha \geq 2$, the results suggest that the setting of $\alpha$ does not significantly influence the ranking of the five stopping criteria.
In contrast, the rankings of the stopping criteria for $\alpha=1$ and $\alpha=2$ are different in some cases.
POSE with $\alpha=1$ does not impose any penalty on stopping criteria in Case 3 discussed in \pref{subsec:pose}.
Thus, the results suggest that the existence of a penalty influences the ranking of stopping criteria based on POSE.





\subsubsection{Effects of $\delta$}

Next, we investigate the effects of $\delta$ in POSE.
\pref{tab:pose_param_delta} shows the Friedman test-based average rankings of the five stopping criteria when setting $\delta$ to $0, 10^{-4}, 10^{-3}, 10^{-2}, $ and $10^{-1}$, where the other settings are the same as in \pref{tab:pose_param_alpha}.
Tables \ref{tab:pose_param_delta_IBEA}--\ref{tab:pose_param_delta_NSGA3} in the supplementary file show the results of the other EMO algorithms.
Here, $\alpha=2$ in this analysis.

As shown in \pref{tab:pose_param_delta}, unlike $\alpha$, the results indicate that the setting of $\delta$ significantly influences the ranking of the five stopping criteria.
For example, as seen from \pref{tab:pose_param_delta}(a), OCD performs the worst for $\delta=0$.
However, the ranking of OCD becomes higher as $\delta$ increases.
In fact, OCD shows the best performance for $\delta=10^{-1}$.
Similar results are observed in Tables \ref{tab:pose_param_delta_IBEA}--\ref{tab:pose_param_delta_NSGA3}.
When using a large $\delta$ value, $\texttt{FE}^{*}$ is small in general.
Thus, POSE with a large $\delta$ value favors stopping criteria that aggressively stop the search once no significant improvement in the best-so-far HV value is observed.


\vspace{0.2em}
\noindent \textit{Summary.} The results show that the ranking of stopping criteria by POSE is not sensitive to the setting of $\alpha$ when $\alpha \geq 2$.
In contrast, the results suggest that the setting of $\delta$ significantly influences the ranking of stopping criteria by POSE.
Based on these observations, as a base case, we recommend $\alpha = 2$ and $\delta = 0$.
After examining the results obtained with the base-case setting, practitioners can investigate the sensitivity of the performance of stopping criteria to $\alpha$ and $\delta$. 
Note that a comparison of stopping criteria based on POSE values is meaningful only when $\alpha$ and $\delta$ are fixed.


\subsection{Effectiveness of the proposed data representation method}
\label{subsec:fsize}

This section investigates the effectiveness of the data representation method described in \pref{subsec:table_ascii}.
As a base line, we use the simple data representation method described in \pref{subsec:txtfile}.


\pref{fig:filesize} shows the results of the two data representation methods on the DTLZ1 problem with $m \in \{2, 4, 6\}$.
For each method, \pref{fig:filesize} shows the amount of memory required to store all objective vectors generated during the search process in text files for 31 runs.
The results of the simple method represent the total file size of $t^{\mathrm{max}}$ text files $\texttt{fP\_1.csv}, $ $\dots, $ $\texttt{fP\_tmax.csv}$ as shown in \pref{fig:example_naive}.
The results of the proposed method represent the total file size of $\texttt{fx.csv}$ and $\texttt{id.csv}$ as shown in \pref{fig:example_naive}.
Note that the results on the other seven problems (DTLZ2, ..., CDTLZ2) are almost the same as that on the DTLZ1 problem shown in \pref{fig:filesize}.
Technically, for both data representation methods, we used the Base64 encoding scheme~\cite{josefsson2006base16} to store real values, which is fully reversible.
Using this encoding scheme slightly reduces the file size.



\begin{figure}[t]
\centering
\includegraphics[width=0.45\textwidth]{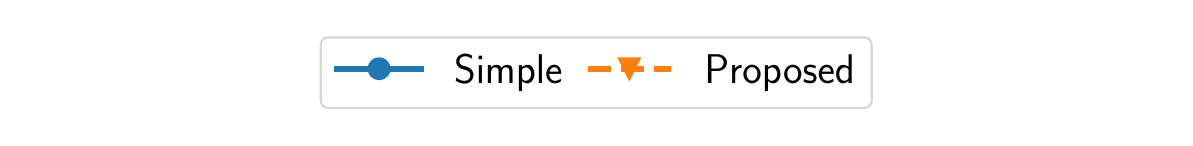}\\[-0.8em]
\subfloat[NSGA-II]{
\includegraphics[width=0.231\textwidth]{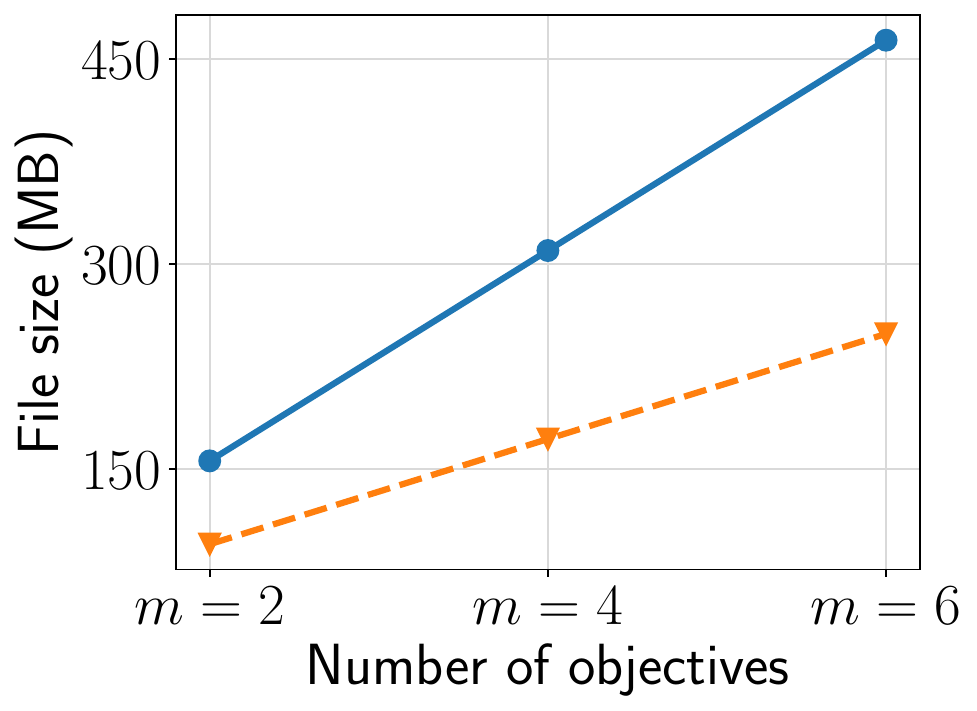}
}
\subfloat[SMS-EMOA]{
\includegraphics[width=0.231\textwidth]{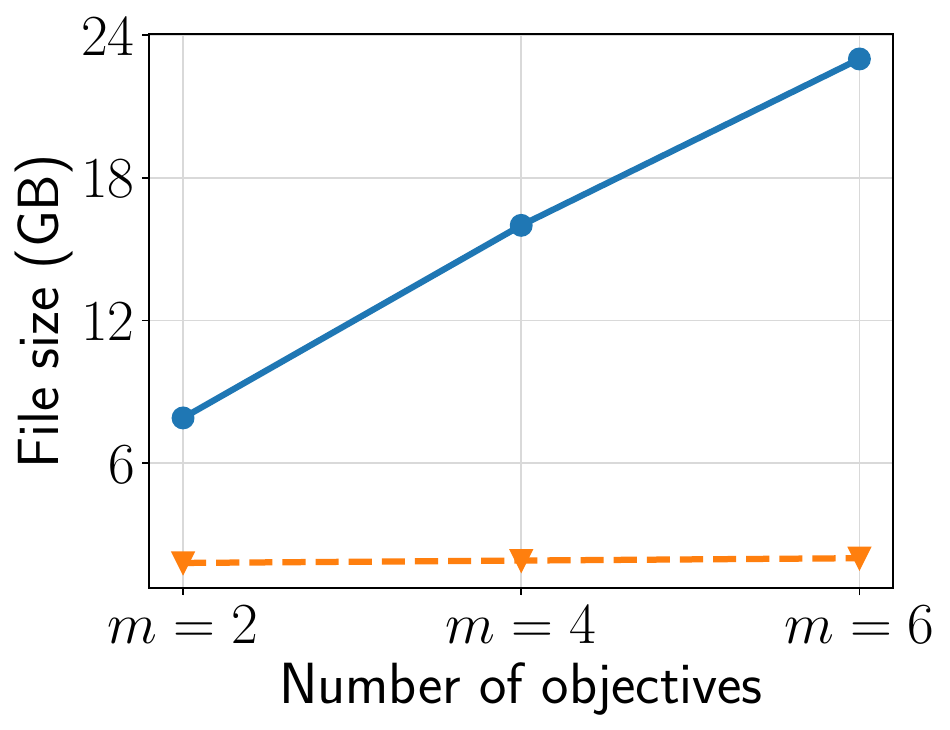}
}\\

\caption{Total size of all text files in the two data representation methods on the DTLZ1 problem with $m \in \{2, 4, 6\}$.}
  \label{fig:filesize}
\end{figure}


Figures \ref{fig:filesize}(a) and (b) show the results of NSGA-II and SMS-EMOA, respectively.
Figure \ref{fig:supp-filesize} in the supplementary file shows the results of IBEA, MOEA/D, and NSGA-III.
We do not describe Figure \ref{fig:supp-filesize} in detail, but it is similar to \pref{fig:filesize}(a).

\pref{fig:filesize}(a) indicates that the proposed method reduces the total file size by over 50 MB compared to the simple method for storing the results of NSGA-II for $m=2$.
The effectiveness of the proposed method becomes more pronounced as $m$ increases.
In fact, the proposed method reduces the file size by about 200 MB compared to the simple method for $m=6$.

As seen from \pref{fig:filesize}(b), the proposed method drastically reduces the file size required to store the results of SMS-EMOA.
Since SMS-EMOA is a steady-state EMO algorithm, the maximum number of iterations in SMS-EMOA is larger than that in NSGA-II. 
For this reason, the effect of the proposed method becomes more significant for SMS-EMOA.
For example, \pref{fig:filesize}(b) shows that the proposed method reduces the total file size by over 20 GB compared to the simple method for $m=6$.


\vspace{0.2em}
\noindent \textit{Summary.} The proposed data representation method can effectively reduce the file size required to store the population states, especially for a large $m$ and steady-state EMO algorithms.

\section{Conclusion}
\label{sec:conclusion}

This paper presented the three contributions, (i)--(iii), towards an effective benchmarking of stopping criteria for EMO.
First, \pref{subsec:pose} proposed (i) POSE, which is a measure for evaluating the performance of stopping criteria for EMO.
Unlike the traditional performance evaluation method, the proposed POSE represents the performance of stopping criteria as a single scalar value while considering (1) the quality of the final population and (2) the budget used in the search.
Then, \pref{subsec:txtfile} proposed (ii) the file-based benchmarking approach that preserves the population states of EMO algorithms in text files, which are then reloaded during benchmarking.
Since the file-based approach does not require reimplementing or rerunning EMO algorithms, it makes the benchmarking process easier.
\pref{subsec:table_ascii} proposed (iii) the data representation method for effectively storing population states in text files.
This method addresses the large-scale storage issue of (ii) the file-based benchmarking approach.



Through experiments, \pref{subsec:issue} demonstrated the difficulty in evaluating the performance of stopping criteria for EMO by using the traditional performance evaluation method.
Then, \pref{subsec:eff_pose} showed that the proposed POSE can enable
a straightforward comparison of stopping criteria.
\pref{subsec:hyperparameter} demonstrated the effects of two parameters $\alpha$ and $\delta$, providing a rule of thumb for setting them.
Finally, \pref{subsec:fsize} found that the proposed data representation method can effectively reduce the file size required to store the population states of EMO algorithms.


Future work should conduct a rigorous benchmarking of stopping criteria for EMO based on our contributions.
Our preliminary results suggest the necessity of parameter tuning to maximize the performance of the five stopping criteria.
Automatic algorithm configuration~\cite{Birattari09} of stopping criteria for EMO is an avenue for future work.
We believe that POSE can evaluate the performance of stopping criteria in single-objective optimizers (e.g., CMA-ES~\cite{Hansen16a}).
Investigating the generality of POSE is left for future work.


\begin{acks}
   This work was supported by JSPS KAKENHI Grant Numbers \seqsplit{25K03194} and \seqsplit{23H00491}.
\end{acks}

\bibliographystyle{ACM-Reference-Format}
\bibliography{reference}

\end{document}